\documentclass{article}

\usepackage[preprint]{neurips_2026}

\usepackage[utf8]{inputenc} 
\usepackage[T1]{fontenc}    
\usepackage{hyperref}       
\usepackage{url}            
\usepackage{booktabs}       
\usepackage{amsfonts}       
\usepackage{nicefrac}       
\usepackage{microtype}      
\usepackage{xcolor}         
\usepackage{subcaption}
\usepackage{graphicx}
\usepackage{amsmath}
\usepackage{amsthm}
\usepackage{threeparttable}
\usepackage{pifont}
\usepackage[dvipsnames]{xcolor}

\theoremstyle{plain}

\theoremstyle{definition}
\newtheorem{definition}{Definition}
\newtheorem{axiom}{Axiom}
\theoremstyle{remark}
\newtheorem{remark}{Remark}

\title{Evaluating the Diversity of AI-Generated Content \\ with Diversity Profiles}

\author{%
  Xiuyuan Hu$^{1,2}$, Xuege Hou$^{1}$, Guoqing Liu$^{3}$, Yang Zhao$^{1}$, Jieran Li$^{1}$, Dongbiao Sun$^{1}$, \\
  \textbf{José Miguel Hernández-Lobato$^{4}$, Hao Zhang$^{1}$\thanks{Corresponding author: Hao Zhang.}~, Xue Liu}$^{2,5,6}$ \\
  $^{1}$Department of Electronic Engineering, Tsinghua University, $^{2}$MBZUAI,\\
  $^{3}$Microsoft Research AI for Science\\
  $^{4}$Department of Engineering, University of Cambridge \\
  $^{5}$McGill University, $^{6}$Mila - Quebec AI Institute \\
  \texttt{huxy22@mails.tsinghua.edu.cn, haozhang@tsinghua.edu.cn} 
}

\begin{document}

\maketitle

\begin{abstract}
Diversity is a fundamental criterion for evaluating generative artificial intelligence (AI) systems, yet its measurement remains inherently ambiguous. Existing approaches typically represent generated samples in an embedding space, compute pairwise distances or similarities, and aggregate them into a single scalar score. Such scalar summaries are convenient, but they often encode different inductive biases and may yield contradictory rankings of the same sample sets. In this paper, we argue that diversity evaluation for AI-generated content is intrinsically under-specified when reduced to a single number. We first review representative diversity metrics, and then diagnose their limitations from two complementary perspectives: an axiomatic analysis showing that no representative scalar metric satisfies all desirable properties simultaneously, and an empirical analysis showing that high-dimensional representation spaces can induce concentrated, modality-dependent distance distributions. To address these issues, we propose \textbf{diversity profiles}: curve-valued, condition-aware summaries that evaluate a parameterized diversity family across a range of thresholds, scales, exponents, or orders under a specified representation and distance or kernel function. Diversity profiles reveal whether a comparison is robust across resolutions or instead depends on an arbitrary parameter choice. We instantiate profiles for several representative metric families and demonstrate their practical use in generative AI evaluation. Overall, diversity profiles provide a more transparent and resolution-aware framework for comparing the diversity of AI-generated content. 
\end{abstract}

\section{Introduction}

Diversity has long been a central concept in the natural sciences. In ecology, it is studied under the name \textit{biodiversity}, where researchers seek to quantify not only how many species are present in a community, but also how evenly individuals are distributed among them and how distinct the species are from one another~\cite{Jost2006}. Classical diversity indices such as Shannon entropy~\cite{ShannonEntropy}, Simpson's index~\cite{SimpsonIndex}, and Hill numbers~\cite{HillNumbers} have therefore played a central role in comparing biological communities. A key lesson from this literature is that diversity is not a single primitive notion: richness, evenness, rarity, and dissimilarity describe related but distinct aspects of a population~\cite{Leinster2012}.

In the era of generative artificial intelligence (AI), diversity evaluation has become an increasingly important and challenging problem. Generative models are expected not only to produce high-quality samples, but also to avoid mode collapse, excessive repetition, and overly narrow coverage of the target domain~\cite{ACL21}. This issue arises across modalities, including images, natural language, graphs, and molecules. In natural language generation, for example, diversity may refer to variation in topics, reasoning paths, lexical choices, syntactic forms, or semantic meanings. These notions operate at different resolutions and are often induced by different representations, making it difficult to summarize diversity by a single universal scalar~\cite{ICML25-Position}.

A common strategy in recent work is to embed generated samples into a representation space, compute pairwise distances or similarities, and then aggregate the resulting matrix into one diversity score. Representative metrics include average pairwise distance~\cite{IntDiv}, energy-based scores~\cite{Energy}, threshold-based packing metrics such as Circles~\cite{Circles}, similarity-spectrum methods such as the Vendi score~\cite{VendiScore,VendiScoreCousins}, and metric-space magnitude~\cite{Magnitude}. These metrics are useful and often interpretable, but they encode different inductive biases. As a result, different metrics can rank the same two generated sets in opposite ways.

This ambiguity is not merely a matter of implementation. We argue that scalar diversity evaluation is intrinsically under-specified. First, widely used diversity metrics satisfy different subsets of desirable axioms, such as size monotonicity, invariance to duplicates, monotonicity with respect to pairwise distances, and continuity. No representative scalar metric satisfies all of these properties simultaneously in its natural domain. Second, many modern diversity metrics contain a scale, threshold, exponent, or order parameter. Different choices of this parameter may lead to different conclusions even within the same metric family. Third, generative AI evaluations are typically performed in high-dimensional representation spaces, where pairwise distances may concentrate and where the meaningful range of scales depends strongly on the embedding model, modality, and distance function.

\begin{figure}[tb]
  \centering
  \begin{subfigure}{0.325\textwidth}
    \centering
    \includegraphics[width=\linewidth]{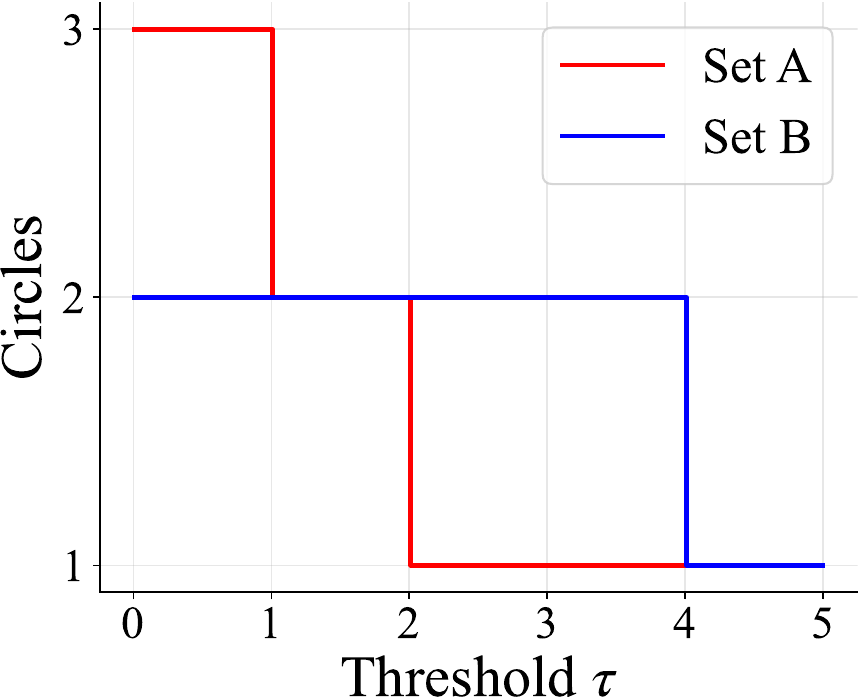}
    \caption{Circles profile}
  \end{subfigure}
  \hfill
  \begin{subfigure}{0.325\textwidth}
    \centering
    \includegraphics[width=\linewidth]{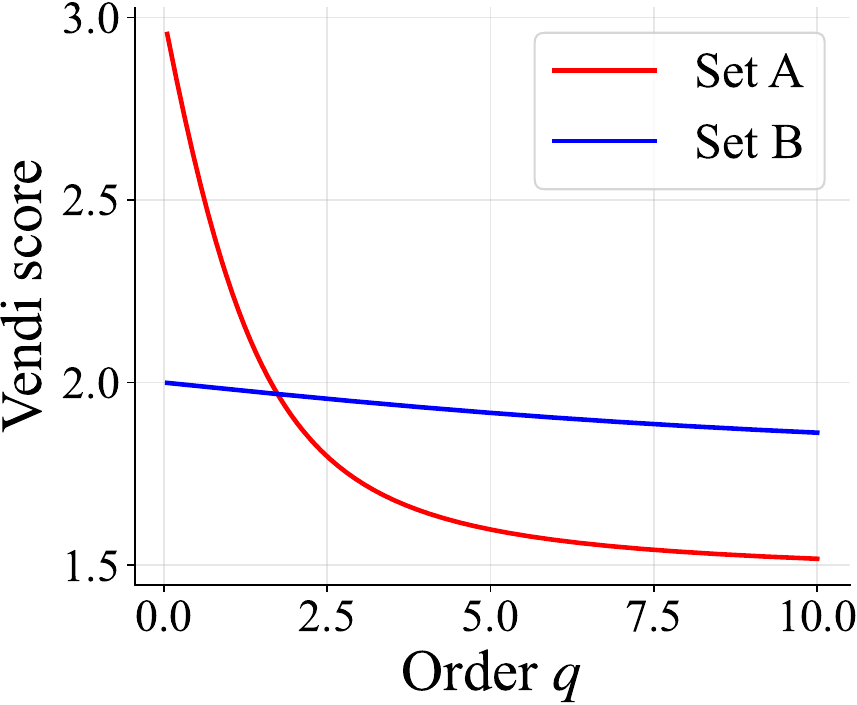}
    \caption{Vendi score profile}
  \end{subfigure}
  \hfill
  \begin{subfigure}{0.325\textwidth}
    \centering
    \includegraphics[width=\linewidth]{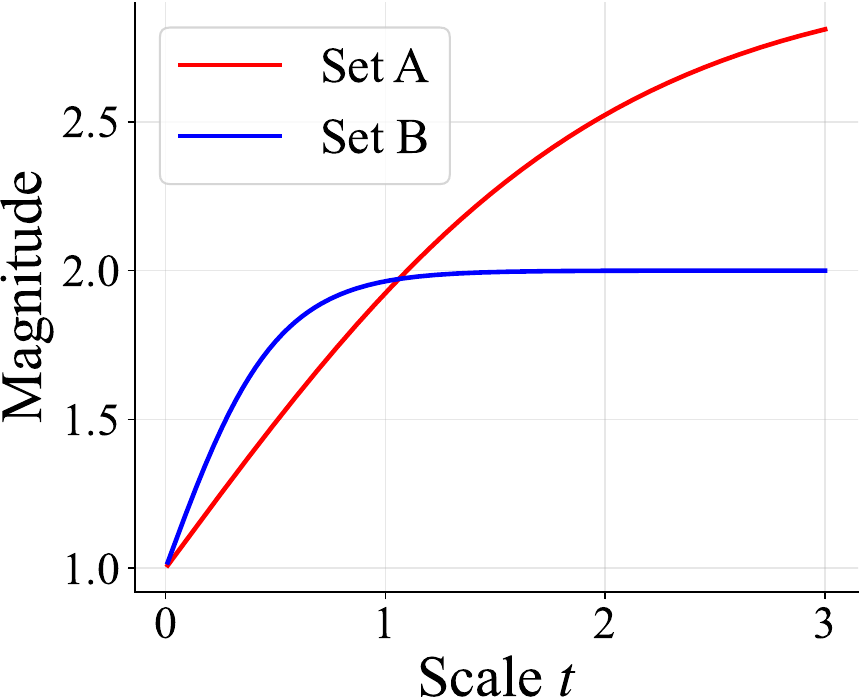}
    \caption{Magnitude profile}
  \end{subfigure}
  \caption{Diversity profiles of a minimal one-dimensional example. Set $A=\{1,2,3\}$, set $B=\{0,4\}$, and distance is measured by absolute difference. Sweeping the profile parameter of three representative diversity metrics reveals resolution-dependent comparisons, illustrating how a single scalar diversity score can obscure important trade-offs.}
  \label{fig:profile_example}
\end{figure}

Figure~\ref{fig:profile_example} illustrates this issue in a minimal example. Richness prefers A, because A contains three distinct elements while B contains two. Average distance prefers B, because the two elements of B are farther apart. Parameterized metrics further reveal that the comparison can depend on the chosen resolution: the Circles, Vendi score, and Magnitude curves may cross as their threshold, order, or scale parameter varies. Thus, reporting a single number can hide important trade-offs and may allow arbitrary metric or parameter choices to determine the claimed conclusion.

To address this problem, we propose to evaluate the diversity of AI-generated content using \textbf{diversity profiles}. Instead of introducing another scalar metric, a diversity profile reports the behavior of a parameterized diversity family across a range of meaningful parameter values. The resulting curve provides a condition-aware summary of diversity under a specified representation, distance or similarity function, metric family, and parameter domain. Diversity profiles are intuitive to interpret: when one curve dominates another across the full parameter domain, the diversity comparison is robust to the choice of resolution; when curves cross, the crossing directly reveals a scale-dependent trade-off that would be obscured by a single scalar score.

Diversity profiles also make it possible to compare generated sample sets not only across resolutions within a fixed metric family, but also across different diversity metric families. Since different metrics emphasize different aspects of diversity, jointly inspecting their profiles helps distinguish conclusions that are metric-family-specific from those that are robust across multiple notions of diversity. Computationally, the additional cost of constructing profiles is modest, making the approach practical for routine generative AI evaluation.

Our contributions are as follows:
\begin{itemize}
    \item We provide a comprehensive review of representative diversity metrics for AI-generated
    content, and diagnose limitations of diversity evaluation with scalar metrics through both an
    axiomatic analysis and the geometry of high-dimensional representation spaces.
    \item We introduce diversity profiles as curve-valued, condition-aware summaries that compare
    generated sample sets across thresholds, scales, exponents, or orders, rather than at a single
    arbitrary parameter value.
    \item We instantiate diversity profiles for several representative metric families, including Energy,
    Circles, Vendi score, and Magnitude, and show how profile comparisons can be used both within
    a metric family across resolutions and across metric families to obtain more robust diversity
    assessments. We also discuss their profile-level properties, intuitive interpretation, and
    computational practicality for generative AI evaluation.
\end{itemize}

Overall, diversity profiles shift the goal of diversity evaluation from selecting a single supposedly best
scalar metric to characterizing how diversity comparisons behave across resolutions and across
metric families. This perspective makes diversity evaluation more transparent, intuitive, less sensitive
to arbitrary hyperparameter choices, and better aligned with the multi-faceted nature of diversity in
generative AI.

\section{Existing Diversity Metrics for AI-Generated Content}
\label{sec:2}

Diversity is a central evaluation criterion for generative AI systems across modalities such as images, text, graphs, and molecules. A common approach is to first represent each generated item in an embedding or feature space, compute pairwise distances between the resulting representations, and then summarize the resulting matrix by a scalar score. We refer to this broad family as \emph{distance-based diversity metrics}. Since the embedding model and the distance function determine the geometry being measured, such metrics should be interpreted as representation-dependent summaries of diversity.

\begin{definition}[Distance-based diversity metric]
Let $X=\{x_1,\dots,x_n\}$ denote a finite collection, or multiset, of generated items. Let $d$ be a nonnegative pairwise distance (dissimilarity) defined on the representation space, and let $D\in\mathbb{R}_{\ge0}^{n\times n}$ be the corresponding distance matrix with entries
\[
D_{ij}=d(x_i,x_j).
\]
A distance-based diversity metric is a permutation-invariant functional
\begin{equation}
\mu_n:\mathcal{D}_n\rightarrow\mathbb{R},
\qquad
\mu_n(D)=\mu_n(PDP^\top)
\end{equation}
for any permutation matrix $P$, where $\mathcal{D}_n$ denotes the admissible set of pairwise distance matrices. Such metrics quantify diversity using only pairwise relations among the elements of $X$.
\end{definition}

We review several representative examples below.

\begin{definition}[Average distance \cite{IntDiv}]
For $n\ge2$, the average pairwise distance (AvgDist) is
\begin{equation}
\mu_{\mathrm{AvgDist}}(D)
:=
\frac{1}{n(n-1)}\sum_{i\ne j}D_{ij}.
\end{equation}
This metric, also commonly referred to as internal diversity, is simple, interpretable, and widely used across generative modeling applications.
\end{definition}

\begin{definition}[Energy \cite{Energy}]
For $s>0$, the energy-based diversity score is
\begin{equation}
\mu_{\mathrm{Energy}}(D;s)
:=
-\frac{1}{n(n-1)}
\sum_{i\ne j}\frac{1}{D_{ij}^s}.
\end{equation}
The exponent $s$ controls the penalty assigned to small pairwise distances. This functional is inspired by the potential energy of repulsive particles: configurations with very close pairs receive large penalties. The score is well defined only when $D_{ij}>0$ for all $i\ne j$; in practice, implementations may add a small $\varepsilon>0$ to avoid singularities.
\end{definition}

\begin{definition}[Circles \cite{Circles}]
For a threshold $\tau\ge0$, define
\begin{equation}
\mu_{\mathrm{Circles}}(D;\tau)
:=
\max_{S\subseteq X}|S|, \text{s.t.}\quad
D_{ij}>\tau,\quad
\forall\, i\ne j \text{ such that } x_i,x_j\in S .
\end{equation}
This metric measures the largest subset of generated items whose elements are mutually separated by more than $\tau$. When $d$ is a metric, it coincides with the packing number~\cite{HighDimProb}.
\end{definition}

\begin{definition}[Vendi score \cite{VendiScore,VendiScoreCousins}]
The Vendi score is similarity-based. Let $K\in\mathbb{R}^{n\times n}$ be a symmetric positive semidefinite similarity matrix with normalized diagonal entries, e.g., $K_{ii}=1$, obtained either directly from a similarity function or indirectly from $D$ through a kernel transformation. Let $\lambda_1,\dots,\lambda_n$ denote the eigenvalues of $K/n$, so that $\lambda_i\ge0$ and $\sum_i\lambda_i=1$. For order $q>0$ and $q\ne1$, the order-$q$ Vendi score is
\begin{equation}
\mu_{\mathrm{Vendi}}(K;q)
:=
\left(\sum_{i=1}^n \lambda_i^q\right)^{\frac{1}{1-q}} .
\end{equation}
The original Vendi score is obtained in the limit $q\to1$:
\begin{equation}
\mu_{\mathrm{Vendi}}(K;1)
:=
\exp\left(-\sum_{i=1}^n \lambda_i\log\lambda_i\right),
\end{equation}
with the convention $0\log 0=0$. Thus, the Vendi score is the exponential of the Shannon entropy when $q=1$ and of the R\'enyi entropy for $q\ne1$. The order-$2$ case is closely related to R\'enyi kernel entropy~\cite{RKE}, and Fourier-feature approximations have been proposed to scale related kernel-entropy computations \cite{FKEA}.
\end{definition}

\begin{definition}[Magnitude \cite{Magnitude}]
Let $Z_t(D)\in\mathbb{R}^{n\times n}$ be the similarity matrix
\[
[Z_t(D)]_{ij}=\exp(-t\cdot D_{ij}),
\qquad t>0 .
\]
If $Z_t(D)$ is nonsingular, the magnitude at scale $t$ is
\begin{equation}
\mu_{\mathrm{Mag}}(D;t)
:=
\mathbf{1}^\top Z_t(D)^{-1}\mathbf{1}
=
\sum_{i,j}[Z_t(D)^{-1}]_{ij}.
\end{equation}
This construction is the finite-space version of magnitude in metric geometry~\cite{Magnitude2} and can be interpreted as an effective size of the metric space at scale $t$.
\end{definition}

\paragraph{Other metrics.}
The above list is not exhaustive. Other recent distance-based proposals include Hamiltonian diversity~\cite{HamDiv}, NovelSum~\cite{NovelSum}, and additional axiomatic constructions~\cite{ICML25-Axioms}. 

Beyond distance-based metrics, diversity is also quantified by \textit{count-based metrics}. A representative example is \textbf{Richness}, defined as the number of unique elements in a batch. 
Domain-specific variants include counting unique words or n-grams in natural language generation and unique molecular scaffolds in cheminformatics. These metrics are robust and easy to interpret, but they do not capture graded distances among distinct generated items.

\section{Diagnosis of Existing Metrics}

\subsection{An Axiomatic Analysis}
\label{sec:3-1}

A useful way to evaluate diversity metrics is through axioms: one specifies desirable properties of a diversity measure and then checks which metrics satisfy them. This perspective appears in several recent studies, although the exact axioms differ across papers \cite{SIGIR18-Axioms,AxiomsBook,ICML25-Axioms}. Here we summarize four desiderata that are particularly relevant for evaluating generated samples. The notation follows Section~\ref{sec:2}, and we write $\mu_n$ when the sample size matters.

\begin{axiom}[Size Monotonicity]
Adding a new element should not decrease diversity. Formally, if $D'\in\mathcal{D}_{n+1}$ extends $D\in\mathcal{D}_n$ by adding one row and column corresponding to a new element $x_{n+1}$, then
\begin{equation}
\mu_{n+1}(D')\ge \mu_n(D).
\end{equation}
\end{axiom}

\begin{axiom}[Twin Property]
Adding a duplicate element should not change diversity. If the added element is a duplicate of some existing element $x_i$, meaning that its distances to all existing elements match those of $x_i$ and $D'_{n+1,i}=0$, then
\begin{equation}
\mu_{n+1}(D')=\mu_n(D).
\end{equation}
This property prevents repeated samples from artificially increasing the diversity score.
\end{axiom}

\begin{axiom}[Distance Monotonicity]
For two admissible distance matrices $D,D'\in\mathcal{D}_n$ of the same size, if
\[
D'_{ij}\ge D_{ij},\qquad \forall\, i\ne j,
\]
then
\begin{equation}
\mu_n(D')\ge \mu_n(D).
\end{equation}
This axiom states that increasing pairwise separations should not reduce measured diversity.
\end{axiom}

\begin{axiom}[Continuity]
The metric should vary continuously under small perturbations of the pairwise distances. For example, under the Frobenius norm,
\begin{equation}
D^{(m)}\rightarrow D
\quad\Rightarrow\quad
\mu_n(D^{(m)})\rightarrow \mu_n(D).
\end{equation}
\end{axiom}

\begin{table}[tbp]
\centering
\caption{Axiomatic analysis of representative diversity metrics.
Symbols: \textcolor{Green}{\ding{51}} = satisfies, \textcolor{red}{\ding{55}} = violates, -- = the metric is not distance-based.}
\label{tab:MetricsAxioms}
\begin{tabular}{l|cccc}
\toprule
\textbf{Diversity Metric}
& \textbf{A1 Size Mono.}
& \textbf{A2 Twin}
& \textbf{A3 Dist. Mono.}
& \textbf{A4 Cont.} \\
\midrule
Richness
& \textcolor{Green}{\ding{51}}
& \textcolor{Green}{\ding{51}}
& --
& -- \\

AvgDist
& \textcolor{red}{\ding{55}}
& \textcolor{red}{\ding{55}}
& \textcolor{Green}{\ding{51}}
& \textcolor{Green}{\ding{51}} \\

Energy ($s$)
& \textcolor{red}{\ding{55}}
& \textcolor{red}{\ding{55}}
& \textcolor{Green}{\ding{51}}
& \textcolor{Green}{\ding{51}} \\

Circles ($\tau$)
& \textcolor{Green}{\ding{51}}
& \textcolor{Green}{\ding{51}}
& \textcolor{Green}{\ding{51}}
& \textcolor{red}{\ding{55}} \\

Vendi score ($q$)
& \textcolor{red}{\ding{55}}
& \textcolor{red}{\ding{55}}
& \textcolor{red}{\ding{55}}
& \textcolor{Green}{\ding{51}} \\

Magnitude ($t$)
& \textcolor{Green}{\ding{51}}
& \textcolor{red}{\ding{55}}
& \textcolor{red}{\ding{55}}
& \textcolor{Green}{\ding{51}} \\

\bottomrule
\end{tabular}
\end{table}

Table~\ref{tab:MetricsAxioms} summarizes the behavior of the representative metrics, with proofs in Appendix~\ref{app:proofs_table1}. No single metric in this list satisfies all four axioms in its natural domain. These incompatibilities suggest that \emph{a single scalar score may not provide a fully reliable description of diversity in all settings}.

\subsection{High-dimensional Representation Spaces}
\label{sec:3-2}

A second challenge arises from the geometry of high-dimensional representation spaces. High-dimensional spaces exhibit behaviors that differ sharply from low-dimensional intuition, a collection of phenomena often referred to as the \emph{curse of dimensionality}~\cite{CurseDim}. One particularly relevant effect is \textbf{distance concentration}: as dimension grows, pairwise distances among random points can become nearly indistinguishable~\cite{BeyerNN,JMLR2018}.

A common asymptotic formulation is the following. Let $P_d$ be a sequence of distributions on $\mathbb{R}^d$, and let $x_1,\dots,x_n\sim P_d$ be i.i.d.\ samples for fixed $n$. Let $\mathrm{dist}_{\max}(d)$ and $\mathrm{dist}_{\min}(d)$ denote the maximum and minimum pairwise distances among these samples. Under standard concentration conditions on the pairwise distance random variable, the relative contrast satisfies
\begin{equation}
\frac{
\mathrm{dist}_{\max}(d)-\mathrm{dist}_{\min}(d)
}{
\mathrm{dist}_{\min}(d)
}
\xrightarrow[d\to\infty]{p}
0 .
\end{equation}
Equivalently, the nearest and farthest pairwise distances become similar in relative terms. This phenomenon weakens distance-based notions such as nearest neighbors, density, locality, and separation, which underlie many scalar diversity metrics~\cite{HighDimProb2,HighDimSurvey2012}.

\begin{figure}[tb]
  \centering
  \begin{subfigure}{0.325\textwidth}
    \centering
    \includegraphics[width=\linewidth]{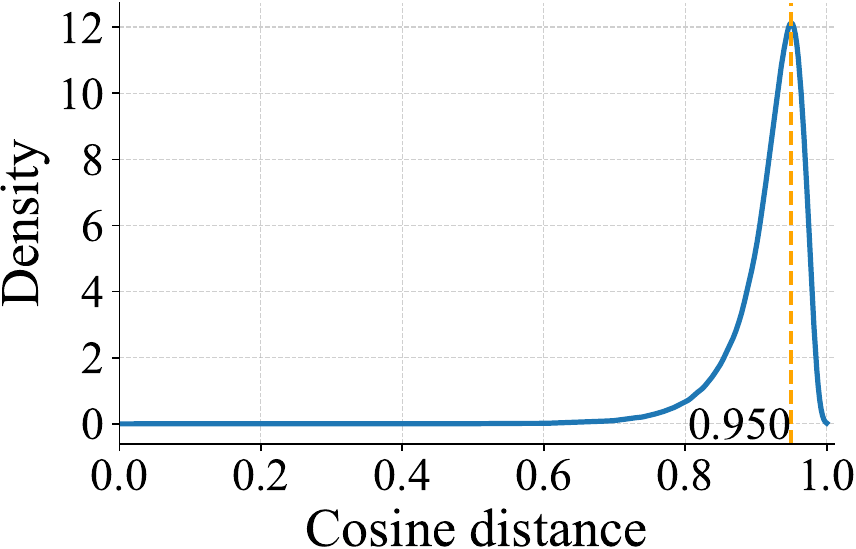}
    \caption{Images}
  \end{subfigure}
  \hfill
  \begin{subfigure}{0.325\textwidth}
    \centering
    \includegraphics[width=\linewidth]{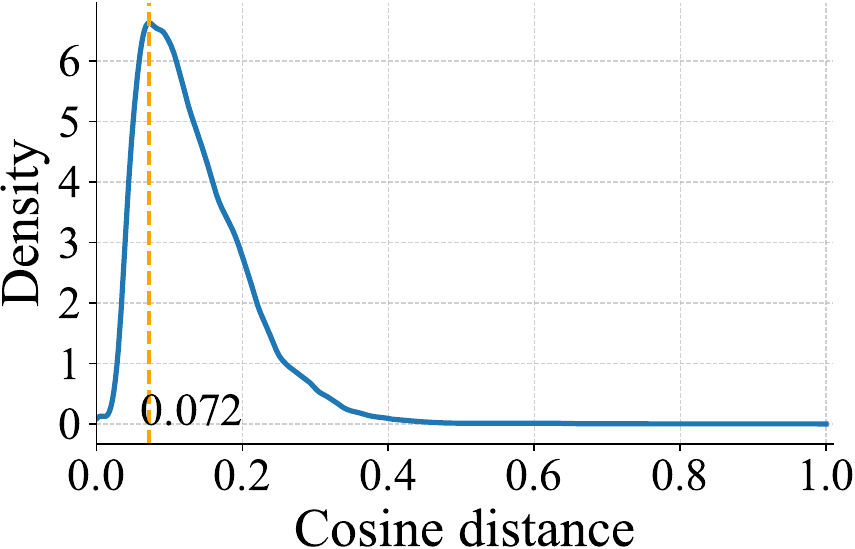}
    \caption{Text}
  \end{subfigure}
  \hfill
  \begin{subfigure}{0.325\textwidth}
    \centering
    \includegraphics[width=\linewidth]{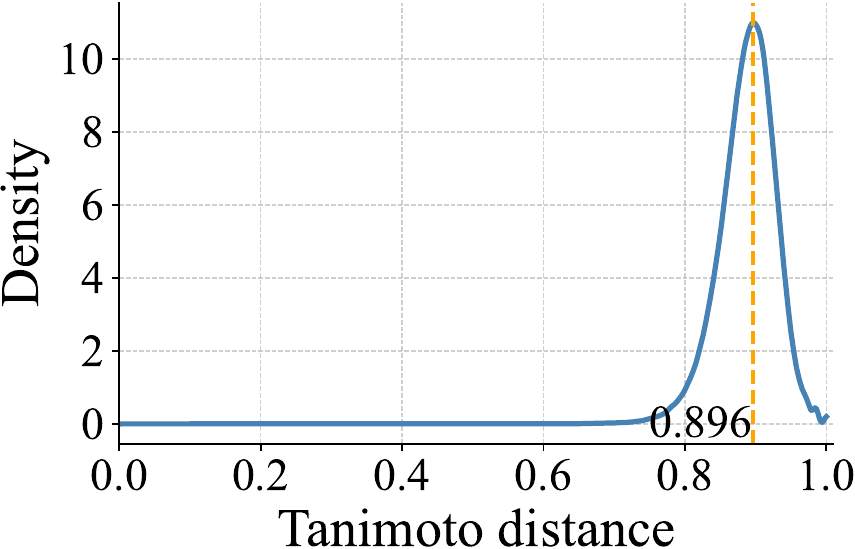}
    \caption{Molecules}
  \end{subfigure}
  \caption{Empirical distributions of pairwise distances in three generative AI domains.}
  \label{fig:ML_dists}
\end{figure}

In generative AI, high dimensionality is not merely a theoretical concern. Common deep representations often have hundreds or thousands of dimensions before any downstream distance is computed. To illustrate this issue, Figure~\ref{fig:ML_dists} reports empirical distributions of pairwise distances in three representative domains: images, natural language text, and molecular structures. The observed concentration across these domains is consistent with high-dimensional representation effects under the chosen embeddings and distance functions. At the same time, the location and shape of the concentrated distributions differ substantially across modalities and distance definitions. These observations sharpen the parameter-selection problem for diversity metrics: when pairwise distances occupy only a narrow empirical range, large regions of the parameter space may become saturated or uninformative, making any single threshold, scale, exponent, or order intrinsically fragile and representation-dependent. This motivates \emph{a more comprehensive, condition-aware view of diversity, rather than reliance on a single absolute score}.
\section{Diversity Profiles}

\subsection{Motivation for Diversity Profiles}

Sections~\ref{sec:3-1} and~\ref{sec:3-2} show that diversity evaluation for generated content is inherently multi-faceted. From a theoretical perspective, no scalar metric considered in Section~\ref{sec:2} satisfies all four desiderata in Table~\ref{tab:MetricsAxioms}: size monotonicity, twin property, distance monotonicity, and continuity. Hence, a single scalar score cannot simultaneously capture all desirable aspects of diversity. Moreover, even within a fixed metric family, the conclusion may depend on the choice of a scale, threshold, or order parameter. As illustrated in Figure~\ref{fig:profile_example}, two sample sets can be ranked differently by the same diversity family under different hyperparameter values.

A closely related idea has a long history in ecology, where \emph{diversity profiles} are used to compare biological communities across the full range of Hill-number orders rather than reporting a single diversity index~\cite{DivProfile,Leinster2012}. In that setting, the order parameter determines the sensitivity of the diversity measure to rare species: small orders emphasize richness and rare types, whereas large orders emphasize dominant types and evenness among common species. This perspective is especially useful when no single order is universally appropriate.

\begin{remark}
Hill numbers, R\'enyi entropy, Shannon entropy, and the Vendi score are closely connected. In particular, Hill numbers are exponentials of R\'enyi entropies, with Shannon entropy recovered in the limit as the order tends to one. The Vendi score applies the same effective-number principle to the spectrum of a normalized similarity matrix. Thus, diversity profiles can be naturally generalized to the Vendi score with an order parameter.
\end{remark}

From a practical perspective, diversity evaluation is condition-specific. A distance-based score depends not only on the generated samples, but also on the representation, distance function, and parameter domain. This dependence becomes especially pronounced in high-dimensional embedding spaces, where pairwise distances often concentrate and where the concentration location and spread vary across modalities, embedding models, and distance functions, as shown in Section~\ref{sec:3-2}. Consequently, a fixed numerical threshold or scale has no universal meaning: it may be too small to distinguish local structure in one representation, too large to avoid degeneracy in another, or sensitive to small changes in conditions. Parameter selection is therefore part of the evaluation problem itself. Diversity profiles can address this by replacing a single parameter choice with a curve over an empirically meaningful range.

This motivates two essential components of a diversity profile. First, the \emph{curve} is important because it reports the behavior of a metric family over a domain of scales, thresholds, or orders, rather than committing to one hyperparameter value. Second, the \emph{condition} is important because the curve is only interpretable relative to the chosen representation, distance, and parameter domain. In this sense, a diversity profile is a condition-aware summary that exposes how diversity comparisons change across resolutions under a specified evaluation setting.

\subsection{Definition and Properties of Diversity Profiles}

Let $X=\{x_1,\ldots,x_n\}$ be a finite multiset of generated samples. A representation map $\phi$ sends each sample to a feature space, and a distance function $d$ induces the pairwise distance matrix
\[
    D_{ij}=d(\phi(x_i),\phi(x_j)).
\]
For similarity-based metrics such as the Vendi score, we analogously use a positive semidefinite similarity matrix $K$ obtained either directly from a similarity function or from $D$ through a fixed kernel transformation.

\begin{definition}[Diversity profile]
A diversity profile for generated content is specified by a tuple
\[
\mathcal{P}=(\phi,d,\mu,\Theta), 
\]
where $\phi$ is the representation, $d$ is the distance function, $\mu$ is a parameterized family of diversity functionals, and $\Theta$ is the admissible set of scale, threshold, or order parameters. For a sample set $X$, the associated profile is the function
\begin{equation}
P_X^{\mathcal{P}}(\theta)=\mu_\theta(D_X), \qquad \theta\in\Theta,
\end{equation}
or $P_X^{\mathcal{P}}(\theta)=\mu_\theta(K_X)$ for similarity-based families.
\end{definition}

Thus, a diversity profile is not a new scalar metric, but a curve-valued summary of how a metric family behaves across resolutions. Richness and Average distance have no intrinsic controlling hyperparameter, so they can be regarded as degenerate, scale-rigid profiles. In contrast, Energy, Circles, Vendi score, and Magnitude naturally define nontrivial profiles.

Table~\ref{tab:profile_properties} summarizes the main profile-level properties of these four families. All four profiles are permutation-invariant because they depend only on the pairwise distance or similarity matrix, not on the ordering of samples. Their behavior with respect to the profile parameter, however, differs substantially.

\begin{table}[t]
\centering
\caption{Profile-level properties of representative diversity families.}
\label{tab:profile_properties}
\begin{tabular}{lccc}
\toprule
Metric family & Parameter & Monotonicity in parameter & Continuity \\
\midrule
Energy & $s>0$ 
& case-dependent 
& continuous \\
Circles & $\tau\ge 0$ 
& nonincreasing 
& stepwise  \\
Vendi score & $q>0$ 
& nonincreasing 
& continuous \\
Magnitude & $t>0$ 
& typically nondecreasing
& continuous \\
\bottomrule
\end{tabular}
\end{table}

\paragraph{Energy profile.}
The exponent $s$ controls the penalty assigned to small distances. The profile is continuous in $s$ whenever all pairwise distances are strictly positive, while its monotonicity depends on the values of distances (compared to $1$).

\paragraph{Circles profile.}
The threshold $\tau$ specifies the minimum separation required between selected samples. As $\tau$ increases, the feasibility constraint becomes stricter, so the profile is nonincreasing. It is an integer-valued step function whose jumps occur at observed pairwise distances. Thus, Circles is particularly interpretable as a resolution-dependent packing number: small thresholds reveal near-duplicate behavior, while large thresholds measure coverage by well-separated representatives.

\paragraph{Vendi score profile.}
The profile is continuous in the order $q$, including near $q=1$. It is nonincreasing in $q$: small $q$ values are sensitive to low-mass spectral components and therefore capture rare directions of variation, whereas large $q$ values emphasize dominant components and measure whether diversity is concentrated in only a few modes.

\paragraph{Magnitude profile.}
The scale $t$ controls the resolution at which points are distinguished. In standard finite metric settings, the profile is continuous and nondecreasing in $t$: small $t$ makes most points highly similar, so the space is viewed coarsely; large $t$ makes only very close points similar, so the space is viewed at a finer resolution.

\subsection{Practical Usage in Generative AI}
\label{sec:4-3}

\begin{figure}
    \centering
    \includegraphics[width=\linewidth]{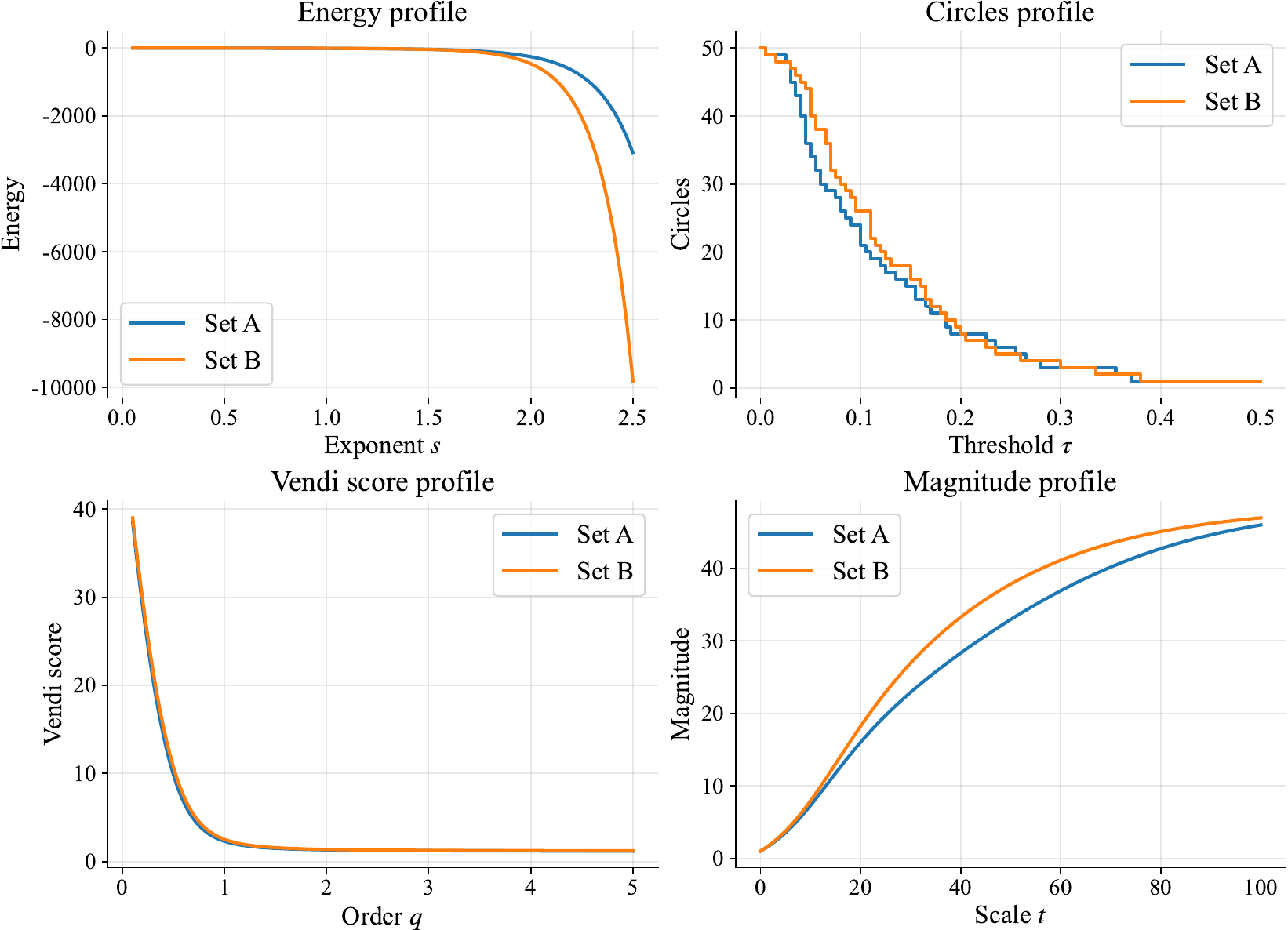}
    \caption{Diversity profiles of two real natural-language sample sets.}
    \label{fig:profile_example2}
\end{figure}

Figure~\ref{fig:profile_example2} gives an example of diversity profiles for two real natural-language sample sets. We use question-answer pairs from the 2WikiMultiHopQA dataset~\cite{2Wiki} and embed them with the Qwen3-8B model~\cite{Qwen3}, which produces $4096$-dimensional dense representations. Each set contains $50$ randomly sampled items. Each curve is evaluated at more than $100$ parameter values, and computing each profile takes less than one second on a standard CPU once the pairwise distance or similarity matrix has been constructed. This suggests that diversity profiles are computationally practical for routine generative AI evaluation. Additional examples of diversity profiles across evaluation settings are provided in Appendix~\ref{app:profiles}.

The interpretation of a profile comparison depends on whether the curves exhibit dominance or crossing. If one set has a higher profile value than another set for all $\theta\in\Theta$, then the diversity ordering is robust to the choice of hyperparameter within that metric family. Such uniform dominance provides stronger evidence than a single scalar comparison, because the conclusion does not depend on an arbitrary scale, threshold, or order.

If the curves cross, then there is no unconditional diversity ordering between the two sets under the chosen profile. Instead, the crossing identifies a scale-dependent trade-off. 
For Energy, superiority at small $s$ mainly reflects larger average pairwise separation under weak sensitivity to close pairs, whereas superiority at large $s$ indicates fewer extremely close pairs, since the inverse-power penalty becomes dominated by the smallest distances. 
For Circles, a set with a higher value at small $\tau$ contains more nonduplicate or locally distinct samples, whereas a higher value at large $\tau$ indicates better coverage by mutually well-separated samples. 
For Vendi score, superiority at small $q$ suggests more rare spectral directions, while superiority at large $q$ suggests a more even distribution among dominant modes. 
For Magnitude, differences at small or moderate $t$ reflect coarse-scale separation, while differences at larger $t$ reflect finer-scale resolution among nearby samples. 

In practice, we recommend choosing $\Theta$ according to the empirical distance or similarity distribution. For threshold profiles such as Circles, $\tau$ can be swept over distance quantiles rather than over an arbitrary linear interval. For kernel-based Vendi score profiles and scale-based Magnitude profiles, the parameter can be chosen so that the induced similarities span a nondegenerate domain between nearly identical and clearly separated samples.

Different diversity metric families may also yield contradictory comparisons. For example, in Figure~\ref{fig:profile_example2}, the Energy profile indicates that set A is more diverse, while Magnitude favors set B; the Circles profile exhibits a crossing pattern, and the Vendi score profile shows negligible separation between the two curves. This discrepancy reflects the fact that different metric families emphasize different aspects of diversity, such as close-pair repulsion, packing at a given threshold, spectral effective rank, or scale-dependent effective size. Therefore, we recommend reporting diversity profiles from multiple metric families when possible. Consistent superiority across several families provides substantially stronger evidence of higher diversity than dominance under a single metric family alone.
\section{Conclusion and Discussion}
\label{sec:5}

In this paper, we study the problem of diversity evaluation for AI-generated content from both theoretical and practical perspectives. We first review representative distance-based diversity metrics, and show that these scalar metrics encode different inductive biases. Through an axiomatic analysis, we observe that none of the representative scalar metrics we analyze satisfies all four desiderata simultaneously. Through an empirical analysis of high-dimensional representation spaces, we further show that pairwise distances can concentrate in a modality- and representation-dependent manner, making any single parameter intrinsically fragile. Motivated by these observations, we propose diversity profiles: curve-valued, condition-aware summaries that evaluate a parameterized diversity family over a meaningful parameter domain under a specified representation and distance function. Diversity profiles reveal whether a diversity comparison is robust across resolutions or instead depends on a particular parameter choice, and reporting profiles from multiple metric families provides a more comprehensive view of diversity than relying on a single scalar score.

Several directions remain open. First, the metric families used to construct diversity profiles are themselves not fully satisfactory from an axiomatic perspective. Although diversity profiles reduce the dependence on a single arbitrary parameter and the use of multiple metric families can exploit their complementarity, a deeper theoretical understanding of the relationships among diversity metrics, axioms, and profile-level properties is still needed. 
Second, diversity evaluation for generative AI may benefit from more systematic subset-level analysis. In ecology, $\alpha$ diversity describes diversity within subcommunities, $\beta$ diversity describes variation among subcommunities, and $\gamma$ diversity describes diversity of the overall metacommunity, with classical decompositions such as $\gamma = \alpha \times \beta$ \cite{ABGBiodiv}. Analogous decompositions could also be useful for generative AI evaluation, which may help distinguish local repetition from global coverage and may provide more informative diagnostics for model comparison.

\paragraph{Limitations.}
First, this work focuses on diversity evaluation rather than the joint evaluation of quality and diversity. In generative AI, these two dimensions are tightly coupled: a system can appear diverse by generating low-quality or off-distribution samples, while a high-quality system may still suffer from mode collapse or limited coverage~\cite{ICML25-Position}. Developing evaluation protocols that jointly characterize fidelity, utility, and diversity remains an important challenge. Second, diversity profiles are condition-dependent. Their interpretation depends on the embedding representation, distance or similarity function, metric family, and parameter domain. Although this condition-dependence is made explicit by the profile formulation, practical deployment in specific domains still requires more detailed guidelines for choosing representations, defining meaningful parameter domains, and interpreting profile dominance or crossing patterns.

\paragraph{Broader impacts.}
Moving from scalar diversity metrics to diversity profiles encourages more transparent, resolution-aware, and less hyperparameter-dependent evaluation of generative AI systems. This can help researchers and practitioners detect mode collapse, excessive repetition, narrow coverage, and metric-specific conclusions that would be hidden by a single score. 

\newpage
\bibliographystyle{plain}
\bibliography{ref}

\newpage
\appendix
\appendix

\section{Proofs of Table~\ref{tab:MetricsAxioms}}
\label{app:proofs_table1}

We provide proofs and counterexamples for the axiomatic summary in Table~\ref{tab:MetricsAxioms}. Throughout, all distance matrices are symmetric with zero diagonal. A \textcolor{Green}{\ding{51}} means that the property holds on the natural domain of the corresponding metric. A \textcolor{red}{\ding{55}} means that either the property fails, or the functional is not well defined on the configurations required by the axiom (for example, exact duplicates for singularity-based functionals).

\paragraph{Richness.}
Richness is the number of distinct elements. Adding an element can either create a new equivalence class or join an existing one, so richness satisfies size monotonicity. Adding an exact duplicate joins an existing equivalence class, so the twin property also holds.

\paragraph{Average pairwise distance.}
For \(n\ge 2\),
\[
\mu_{\mathrm{AvgDist}}(D)
=
\frac{1}{n(n-1)}\sum_{i\ne j} D_{ij}.
\]
Distance monotonicity follows immediately from linearity: if
\(D'_{ij}\ge D_{ij}\) for all \(i\ne j\), then
\[
\mu_{\mathrm{AvgDist}}(D')-\mu_{\mathrm{AvgDist}}(D)
=
\frac{1}{n(n-1)}
\sum_{i\ne j}(D'_{ij}-D_{ij})
\ge 0.
\]
Continuity also follows from linearity.

Size monotonicity fails. Consider two points with distance \(1\):
\[
D =
\begin{pmatrix}
0 & 1\\
1 & 0
\end{pmatrix},
\qquad
\mu_{\mathrm{AvgDist}}(D)=1.
\]
Add a duplicate of the first point:
\[
D' =
\begin{pmatrix}
0 & 1 & 0\\
1 & 0 & 1\\
0 & 1 & 0
\end{pmatrix}.
\]
Then
\[
\mu_{\mathrm{AvgDist}}(D')
=
\frac{2(1+0+1)}{3\cdot 2}
=
\frac{2}{3}
< 1.
\]
Thus A1 fails. The same example shows that the twin property fails, since
adding a duplicate changes the score from \(1\) to \(2/3\).

\paragraph{Energy.}
For fixed \(s>0\),
\[
\mu_{\mathrm{Energy}}(D;s)
=
-\frac{1}{n(n-1)}
\sum_{i\ne j}\frac{1}{D_{ij}^s},
\]
on the domain where all off-diagonal distances are strictly positive.

Distance monotonicity holds on this domain. If \(D'_{ij}\ge D_{ij}>0\), then
\[
\frac{1}{(D'_{ij})^s}\le \frac{1}{D_{ij}^s},
\]
and therefore
\[
\mu_{\mathrm{Energy}}(D';s)
\ge
\mu_{\mathrm{Energy}}(D;s).
\]
Continuity follows because \(x\mapsto -x^{-s}\) is continuous on
\((0,\infty)\), and the score is a finite average of such terms.

Size monotonicity fails. Take three points on the real line:
\(x_1=0\), \(x_2=1\), and add \(x_3=\varepsilon\), where
\(0<\varepsilon<1\). For the original two-point set,
\[
\mu_{\mathrm{Energy}}(D;s)=-1.
\]
For the three-point set,
\[
\mu_{\mathrm{Energy}}(D';s)
=
-\frac{1}{3}
\left(
1+\varepsilon^{-s}+(1-\varepsilon)^{-s}
\right),
\]
which is smaller than \(-1\) for sufficiently small \(\varepsilon\). Hence A1
fails.

The twin property also fails in the natural sense that exact duplicates are not
in the finite domain of the functional: adding a duplicate creates a zero
off-diagonal distance, making \(D_{ij}^{-s}\) singular. Thus the score is not a
well-defined real-valued functional on multisets with twins. In
\(\varepsilon\)-regularized implementations, the duplicate contributes a large
negative penalty and therefore changes the score.

\paragraph{Circles.}
For fixed \(\tau\ge 0\),
\[
\mu_{\mathrm{Circles}}(D;\tau)
=
\max_{S\subseteq X}
\left\{
|S| : D_{ij}>\tau \text{ for all distinct } x_i,x_j\in S
\right\}.
\]
Size monotonicity holds because, after adding a new point, every feasible subset
of the original set remains feasible. Hence the maximum cannot decrease.

The twin property holds. Suppose the added point \(x_{n+1}\) is a duplicate of
\(x_i\). Since \(D'_{i,n+1}=0\le \tau\), no feasible subset can contain both
\(x_i\) and \(x_{n+1}\). Moreover, any feasible subset containing \(x_{n+1}\)
but not \(x_i\) can replace \(x_{n+1}\) by \(x_i\), because the duplicate has
the same distances to all other points. Therefore the maximum feasible
cardinality is unchanged.

Distance monotonicity holds. If \(D'_{ij}\ge D_{ij}\) for all \(i\ne j\), then
any subset satisfying \(D_{ij}>\tau\) also satisfies \(D'_{ij}>\tau\). Thus the
family of feasible subsets can only grow, and
\[
\mu_{\mathrm{Circles}}(D';\tau)
\ge
\mu_{\mathrm{Circles}}(D;\tau).
\]

Continuity fails because the functional is thresholded and integer-valued. For
\(n=2\) and \(\tau=1\), let
\[
D^{(m)} =
\begin{pmatrix}
0 & 1+1/m\\
1+1/m & 0
\end{pmatrix},
\qquad
D =
\begin{pmatrix}
0 & 1\\
1 & 0
\end{pmatrix}.
\]
Then \(D^{(m)}\to D\), but
\(\mu_{\mathrm{Circles}}(D^{(m)};\tau)=2\) for all \(m\), whereas
\(\mu_{\mathrm{Circles}}(D;\tau)=1\), because the constraint is strict:
\(D_{12}>\tau\).

\paragraph{Vendi score.}
Let \(K\succeq 0\) have diagonal entries \(K_{ii}=1\), and let
\(\lambda_1,\ldots,\lambda_n\) be the eigenvalues of \(K/n\). For \(q>0\),
\(q\ne 1\),
\[
\mu_{\mathrm{Vendi}}(K;q)
=
\left(\sum_{i=1}^n \lambda_i^q\right)^{1/(1-q)},
\]
with the Shannon limit at \(q=1\).

Continuity holds because the eigenvalues of a symmetric matrix are continuous
functions of the matrix entries, and the maps
\[
(\lambda_i)_i \mapsto
\left(\sum_i \lambda_i^q\right)^{1/(1-q)}
\quad\text{and}\quad
(\lambda_i)_i \mapsto
\exp\left(-\sum_i \lambda_i\log \lambda_i\right)
\]
are continuous on the probability simplex, with the convention
\(0\log 0=0\).

Size monotonicity fails. For two orthogonal items,
\[
K=I_2,
\]
the normalized eigenvalues are \((1/2,1/2)\), so
\(\mu_{\mathrm{Vendi}}(K;q)=2\) for all \(q>0\). Add a duplicate of the first
item:
\[
K'=
\begin{pmatrix}
1 & 0 & 1\\
0 & 1 & 0\\
1 & 0 & 1
\end{pmatrix}.
\]
The eigenvalues of \(K'/3\) are \((2/3,1/3,0)\). Hence
\[
\mu_{\mathrm{Vendi}}(K';q)
=
\left[
(2/3)^q+(1/3)^q
\right]^{1/(1-q)}
<2
\]
for \(q\ne 1\), and the same strict inequality holds at \(q=1\) by the Shannon
limit. Thus adding an element can decrease the score. The same example also
shows that the twin property fails, since the added element is an exact
duplicate but the score changes.

Distance monotonicity is not guaranteed for the Vendi family. To give an
explicit counterexample, take \(q=1/2\) and define distances by
\(D_{ij}=1-K_{ij}\). Let
\[
K=
\begin{pmatrix}
1 & 0.8 & 0.4\\
0.8 & 1 & 0.3\\
0.4 & 0.3 & 1
\end{pmatrix},
\qquad
K'=
\begin{pmatrix}
1 & 0.8 & 0.35\\
0.8 & 1 & 0\\
0.35 & 0 & 1
\end{pmatrix}.
\]
Both matrices are positive semidefinite, and \(K'_{ij}\le K_{ij}\) for all
\(i\ne j\), so the associated distances satisfy \(D'_{ij}\ge D_{ij}\).
However,
\[
\mu_{\mathrm{Vendi}}(K;1/2)\approx 2.5093,
\qquad
\mu_{\mathrm{Vendi}}(K';1/2)\approx 2.4747.
\]
Thus increasing dissimilarities can decrease the Vendi score.

\paragraph{Magnitude.}
For fixed \(t>0\), let
\[
Z_t(D)_{ij}=\exp(-tD_{ij}),
\qquad
\mu_{\mathrm{Mag}}(D;t)=\mathbf{1}^{\top}Z_t(D)^{-1}\mathbf{1}.
\]

Continuity holds on the domain where \(Z_t(D)\) is nonsingular, since
\(D\mapsto Z_t(D)\) is continuous and matrix inversion is continuous on the
open set of nonsingular matrices.

Size monotonicity holds under the standard positive-definite assumption. Let
\(Z\) be the similarity matrix for the original set and suppose the enlarged
matrix has the block form
\[
Z'=
\begin{pmatrix}
Z & z\\
z^\top & 1
\end{pmatrix},
\]
with \(Z'\succ 0\). Let \(w=Z^{-1}\mathbf{1}\) and let
\[
s=1-z^\top Z^{-1}z.
\]
Since \(Z'\succ 0\), the Schur complement satisfies \(s>0\). The block inverse
formula gives
\[
\mathbf{1}^\top (Z')^{-1}\mathbf{1}
=
\mathbf{1}^\top Z^{-1}\mathbf{1}
+
\frac{(1-z^\top w)^2}{s}
\ge
\mathbf{1}^\top Z^{-1}\mathbf{1}.
\]
Therefore magnitude satisfies size monotonicity on this positive-definite
domain.

The twin property fails as a real-valued property on multisets because adding an
exact duplicate makes two rows and columns of \(Z_t(D)\) identical. Hence
\(Z_t(D)\) becomes singular, so the inverse-based magnitude is not defined.

Distance monotonicity fails. Let \(t=1\), and define two similarity matrices
\[
Z=
\begin{pmatrix}
1 & 0.1 & 0.3\\
0.1 & 1 & 0.9\\
0.3 & 0.9 & 1
\end{pmatrix},
\qquad
Z'=
\begin{pmatrix}
1 & 0.1 & 0.1\\
0.1 & 1 & 0.9\\
0.1 & 0.9 & 1
\end{pmatrix}.
\]
Both matrices are positive definite. Let \(D_{ij}=-\log Z_{ij}\) and
\(D'_{ij}=-\log Z'_{ij}\). Since \(Z'_{ij}\le Z_{ij}\) for all \(i\ne j\), we
have \(D'_{ij}\ge D_{ij}\). However,
\[
\mu_{\mathrm{Mag}}(D;1)
=
\mathbf{1}^\top Z^{-1}\mathbf{1}
=
\frac{15}{8}
=
1.875,
\]
whereas
\[
\mu_{\mathrm{Mag}}(D';1)
=
\mathbf{1}^\top (Z')^{-1}\mathbf{1}
=
\frac{175}{94}
\approx 1.8617.
\]
Thus increasing pairwise distances can reduce magnitude.

\paragraph{Remark on the assumptions for magnitude.}
The proof of size monotonicity above uses positive definiteness of the enlarged
similarity matrix. If Definition~6 only assumes nonsingularity, size
monotonicity is not guaranteed. Thus the A1 \textcolor{Green}{\ding{51}} for magnitude should be read as holding under the positive-definite finite-metric/kernel setting, not under mere nonsingularity.

\section{Details of Figure~\ref{fig:ML_dists}}
\label{app:fig_ml_dists}

For each domain in Figure~\ref{fig:ML_dists}, we randomly sample data instances from commonly-used datasets, compute their representations using a standard feature extractor, and estimate the empirical distribution from $100{,}000$ randomly sampled pairwise distances. The goal of this experiment is to illustrate that commonly used high-dimensional representations in different modalities often lead to strongly concentrated pairwise distance distributions.

\paragraph{Images.}
Images are commonly represented by deep neural networks trained for image recognition. We randomly sample RGB images from ImageNet (research-only access terms)~\cite{ImageNet} and extract features using a ResNet-50 encoder (TorchVision, BSD-3-Clause; pretrained weights subject to ImageNet access terms)~\cite{ResNet}. Specifically, each image is mapped to a $2048$-dimensional feature vector from the penultimate representation layer. We then compute pairwise cosine distances between these image embeddings. Cosine distance between deep image features is widely used in image retrieval, clustering, and diversity evaluation. As shown in Figure~\ref{fig:ML_dists}\textit{(a)}, the resulting distance distribution is sharply concentrated.

\paragraph{Natural language text.}
For natural language, modern large language models provide high-dimensional semantic embeddings that are widely used in similarity search, retrieval-augmented generation, and text evaluation. We use question-answer pairs from the 2WikiMultiHopQA dataset (Apache-2.0)~\cite{2Wiki} and embed them with the Qwen3-8B model (Apache-2.0)~\cite{Qwen3}, which produces $4096$-dimensional dense representations. Pairwise dissimilarity is measured by cosine distance. Figure~\ref{fig:ML_dists}\textit{(b)} shows that the empirical distance distribution for textual embeddings is also strongly concentrated.

\paragraph{Molecular structures.}
In cheminformatics, molecular diversity is often assessed using handcrafted molecular fingerprints rather than neural embeddings. We represent molecules sampled from the ChEMBL database (CC BY-SA 3.0)~\cite{ChEMBL} using extended-connectivity fingerprints (ECFP; computed with RDKit, BSD-3-Clause)~\cite{ECFP}, a widely used binary representation that encodes local molecular substructures. We use fixed-length $2048$-dimensional fingerprints and compute pairwise Tanimoto distances~\cite{TanimotoDist}, a standard dissimilarity measure for molecular fingerprints. As illustrated in Figure~\ref{fig:ML_dists}\textit{(c)}, the resulting molecular distances also exhibit strong concentration. This suggests that high-dimensional distance concentration arises not only in learned neural embeddings, but also in classical sparse or binary representations.

Overall, Figure~\ref{fig:ML_dists} demonstrates that distance concentration is a recurring empirical phenomenon across modalities, representation types, and distance functions, and the precise location and shape of the distributions depend on the domain and representation.

\section{More Examples of Diversity Profiles}
\label{app:profiles}

\begin{figure}
    \centering
    \includegraphics[width=\linewidth]{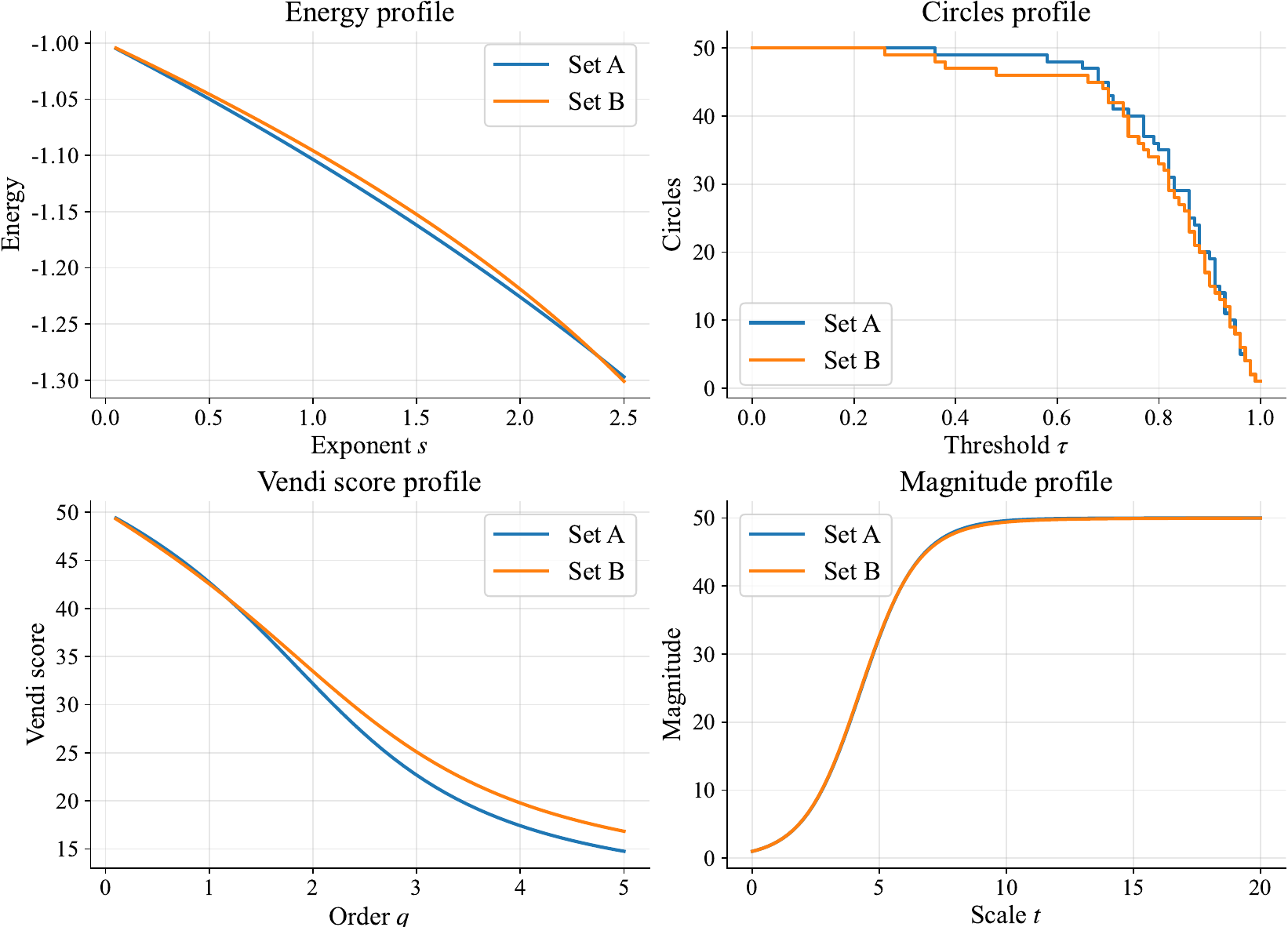}
    \caption{Diversity profiles of two real image sample sets.}
    \label{fig:profile_example3}
\end{figure}

\begin{figure}
    \centering
    \includegraphics[width=\linewidth]{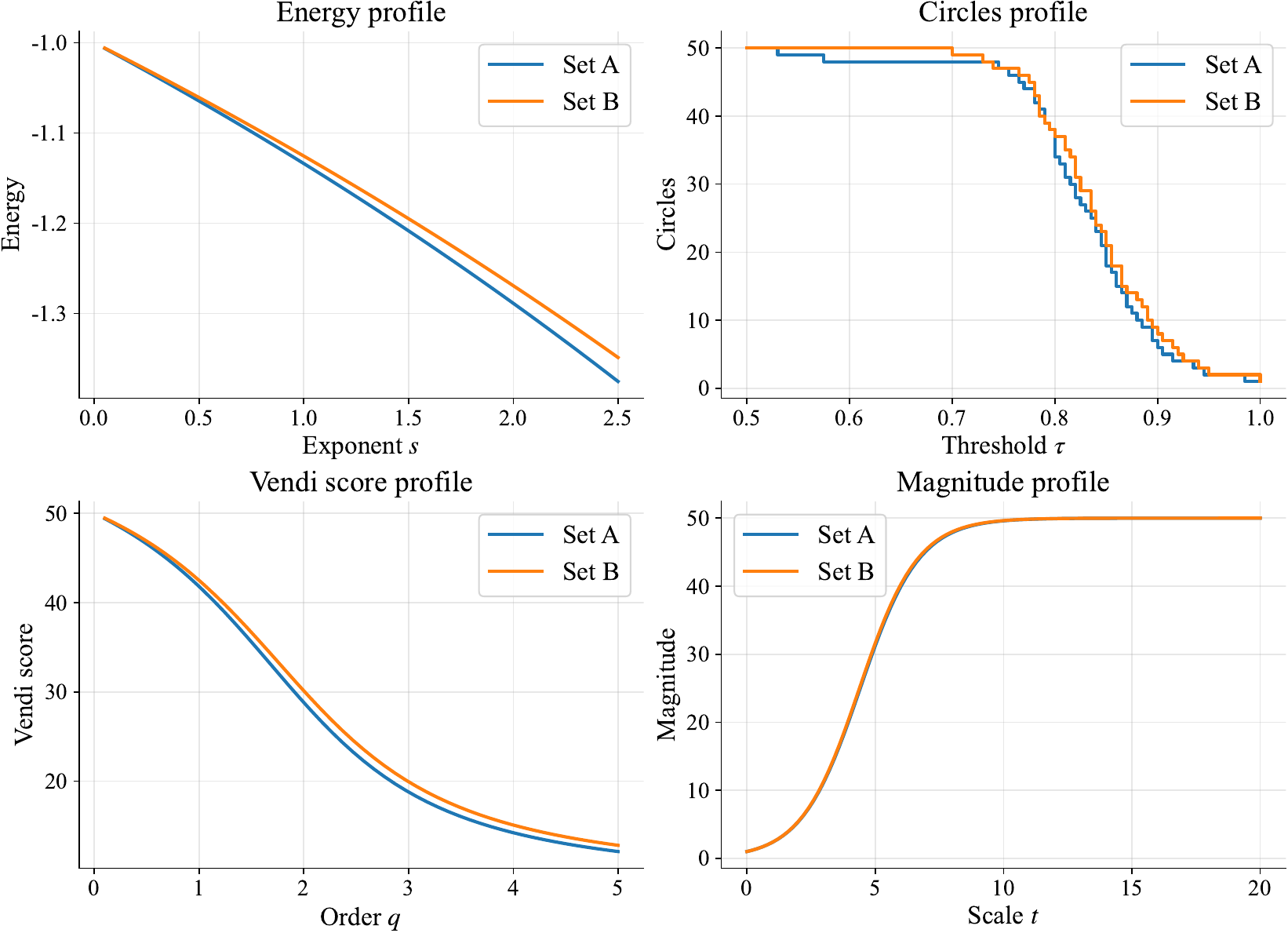}
    \caption{Diversity profiles of two real molecular sample sets.}
    \label{fig:profile_example4}
\end{figure}

In addition to the natural-language example in Figure~\ref{fig:profile_example2}, we provide two further real examples of diversity profiles on different data modalities: images (Figure~\ref{fig:profile_example3}) and molecular structures (Figure~\ref{fig:profile_example4}). For each modality, we use the same data source, representation, and distance function as described in Appendix~\ref{app:fig_ml_dists}. Each of sets A and B consists of 50 data items sampled uniformly at random.

\end{document}